\documentclass{article}[11pt]
\usepackage{epsfig,natbib}

\title{Measuring Latent Causal Structure}
\author{Ricardo Silva \\Department of Statistical Science\\University College London \\ \tt{ricardo@stats.ucl.ac.uk}}

\begin{document}
\maketitle

\section{Introduction}
\label{sec:intro}

Discovering latent representations of the observed world has become
increasingly more relevant in the artificial intelligence literature
\citep{hinton:06,bengio:07}. Much of the effort concentrates on building
latent variables which can be used in prediction problems, such as
classification and regression.

A related goal of learning latent structure from data is that of
identifying which hidden common causes generate the
observations. This becomes relevant in applications that require
predicting the effect of policies.

As an example, consider the problem of identifying the effects of the
``industrialization level'' of a country on its ``democratization
level'' across two different time points. Democratization levels and
industrialization levels are not directly observable: they are hidden
common causes of observable {\it indicators} which can be recorded and
analyzed. For instance, gross national product (GNP) is an indicator
of industrialization level, while expert assessments of freedom of
press can be used as indicators of democratization. Extended discussions
on the distinction between indicators and the latent variables they measure
can be found in the literature of structural equation models \citep{bol:89}
and error-in-variables regression \citep{carroll:95}.

Causal networks can be used as a language to
represent this information. We postulate a graphical encoding of
causal relationships among random variables, where vertices in the
graph representing random variables and directed edges $V_i
\rightarrow V_j$ represent the notion that $V_i$ is a direct cause of
$V_j$. Formal definitions of direct causation and causal networks are
given by \cite{sgs:00} and \cite{pearl:00}.

\begin{figure}[t]
\begin{center}
\epsfig{file=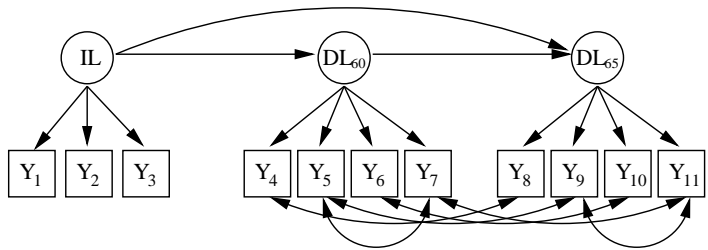,width=9.3cm}
\end{center}
\caption{A causal diagram connecting industrialization levels ($IL$) 
of a country, in 1960, to democratization levels in 1960 and 1965
($DL_{60}$ and $DL_{65}$, respectively). In our diagrams, we follow
the standard structural equation modeling notation and use square
vertices represent observable variables, circles represent latent
variables (Bollen, 1989). The industrialization indicators are: $Y_1$,
gross national product; $Y_2$, energy consumption per capita;
$Y_3$, percentage of labor force in industry. The democratization
indicators are: $Y_4/Y_8$, freedom of press; $Y_5/Y_9$, freedom of opposition;
$Y_6/Y_{10}$, fairness of elections; $Y_7/Y_{11}$, 
elective nature of legislative body. Details are given by Bollen (1989).}
\label{fig:bollen}
\end{figure}

In our setup, we explicitly represent latent variables of interest as
vertices in the graph. For example, in Figure \ref{fig:bollen} we have
a network representation for the problem of causation between
industrialization and democratization levels. This model makes
assumptions about the connections among latent variables themselves:
e.g., industrialization causes democratization, and the possibility of
unmeasured confounding between industrialization and democratization
is not taken into account (which, of course, can be criticized and
refined). 

Following the mixed graph notation
\citep{richardson:02,sgs:00,pearl:00}, we also use bi-directed edges 
$V_i \leftrightarrow V_j$ to denote implicit paths due to latent common
causes. That is, $V_i \leftrightarrow V_j$ denotes a set of causal
paths (e.g., $V_i \leftarrow X \rightarrow V_j$) that originate from
common causes that have been marginalized (such as $X$ in the previous
example), as discussed in full detail by Richardson and Spirtes
\citeyear{richardson:02}. The distinction between ``explicit'' and
``implicit'' latent variables is problem dependent: if we do not wish
to establish causal effects for some hidden variables, then they can
be marginalized.

Establishing the causal connections among latent variables is an
important causal question, but it is only meaningful if such hidden
variables are connected to our observations. A complementary and
perhaps even more fundamental problem is that of finding which latent
variables are out there, and how they cause the observed measures.

This will be the main problem tackled in our contribution: given a
dataset of indicators assumed to be generated by {\it unknown} and
{\it unmeasured} common causes, we wish to discover which hidden common
causes are those, and how they generate our data. Using
a definition from the structural equation modeling literature, we say
we are interested in learning the {\it measurement model} of our
problem \citep{bol:89}. 

In the context of the example of Figure \ref{fig:bollen}, suppose we
are given a dataset with 11 indicators, and wish to unveil the
respective latent common causes and measurement model.  Assuming
Figure \ref{fig:bollen} as the unknown gold standard, we are
successful if we predict that $\{Y_1, Y_2, Y_3\}$ are generated by a
particular hidden common cause, $\{Y_4, \dots, Y_7\}$ are generated by
another hidden common cause and so on. Notice that interpreting the
resulting latent variables and linking them to real entities and
possible interventions requires knowledge of the domain. However,
their existence and relationship to the observations follow from our
data and assumptions, not from a posthoc interpretation.

The solution to this problem lies at the intersection of artificial
intelligence techniques to infer causal structures, statistical models
and the exploitation of assumptions commonly made in some applied
sciences such as psychology and social sciences.

Success will depend on how structured the real-world causal network is
and how valid our assumptions are. If the postulated true network that
generated our data is not sparse, for instance, then there will be so
many models compatible with the observed data that no useful
conclusion can be made. This situation, however, is not different from
the limitations of standard causal network discovery procedures (with no latent
variables) \citep{sgs:00}, which rely on the existence of many
conditional independence constraints.

We describe our assumptions and a formal problem statement in full detail
in Section \ref{sec:statement}. An algorithm to tackle the stated problem
is provided in Section \ref{sec:procedure}. Experiments are show in Section
\ref{sec:experiments}, followed by a Conclusion. Before that, however, we
discuss what is the current common practice for unveiling the causal
measurement structure of the world, and why they fall short on providing
a reasonable solution.

\subsection{A motivating example}

Factor analysis is still the method of choice for suggesting hidden
common cause models in the sciences. A detailed description of the
method within the context of psychology and social sciences is given
by \cite{bart:99}. In this section, we will illustrate the weaknesses
of factor analysis. This motivates the need for more advanced methods
resulting from artificial intelligence techniques in causal discovery.

In a nutshell, the main assumption of factor analysis states that each
observed variable $Y_i$ should be the effect of a set of latent factors
$\mathbf X \equiv \left\{X_1, \dots, X_L\right\}$ plus some
independent error term $\epsilon_i$:

\begin{equation}
\label{eq:linear}
Y_i = \sum_{j = 1}^L \lambda_{ij}X_j + \epsilon_i
\end{equation}

Variables are assumed to be jointly Gaussian, although this is not
strictly necessary. The measurement model is given by the coefficients
$\{\lambda_{ij}\}$ and variances of the error terms $\{\epsilon_i\}$.
Learning the measurement model is the key task, which is required in
order to understand what the hidden common causes should represent in
the real-world. The factor analysis model is agnostic with respect to
the causal structure of $\mathbf X$, but knowing the measurement model
would also help us to learn the causal structure among latent
variables \citep{sgs:00}. In the following discussion, we will assume that we
know how many latent variables exist, and then illustrate how such a
widely used method is unreliable even under this highly favorable
circumstance.

Given the observed covariance matrix of $\mathbf Y \equiv \{Y_1,
\dots, Y_p\}$, it is possible to infer the coefficients 
$\lambda_{ij}$ and the covariance matrix of $\mathbf X$, but not in an
unique way. Without going into details, there are ways of choosing a
solution among this equivalence class such that the measurement
structure is as simple ``as possible'' (within the selection criterion
of choice) \citep{bart:99}. Simplicity here means having many
coefficients $\{\lambda_{ij}\}$ set to zero, indicating that each
observed variable measures only a few of the latent variables. Getting
the correct sparse structure is essential in order to interpret what
the hidden common causes are. Notice that this corresponds to a
directed causal network, where non-zero coefficients are encoded as
directed edges in the graph.

Such methods will work when the true model that generated the data is
in fact a ``simple structure,'' or a ``pure measurement model,'' in the
sense that each observed variable has a single parent in the
corresponding causal network. However, any deviance from this simple
structure will strongly compromise the result.

\begin{figure}[t]
\begin{tabular}{cc}
\epsfig{file=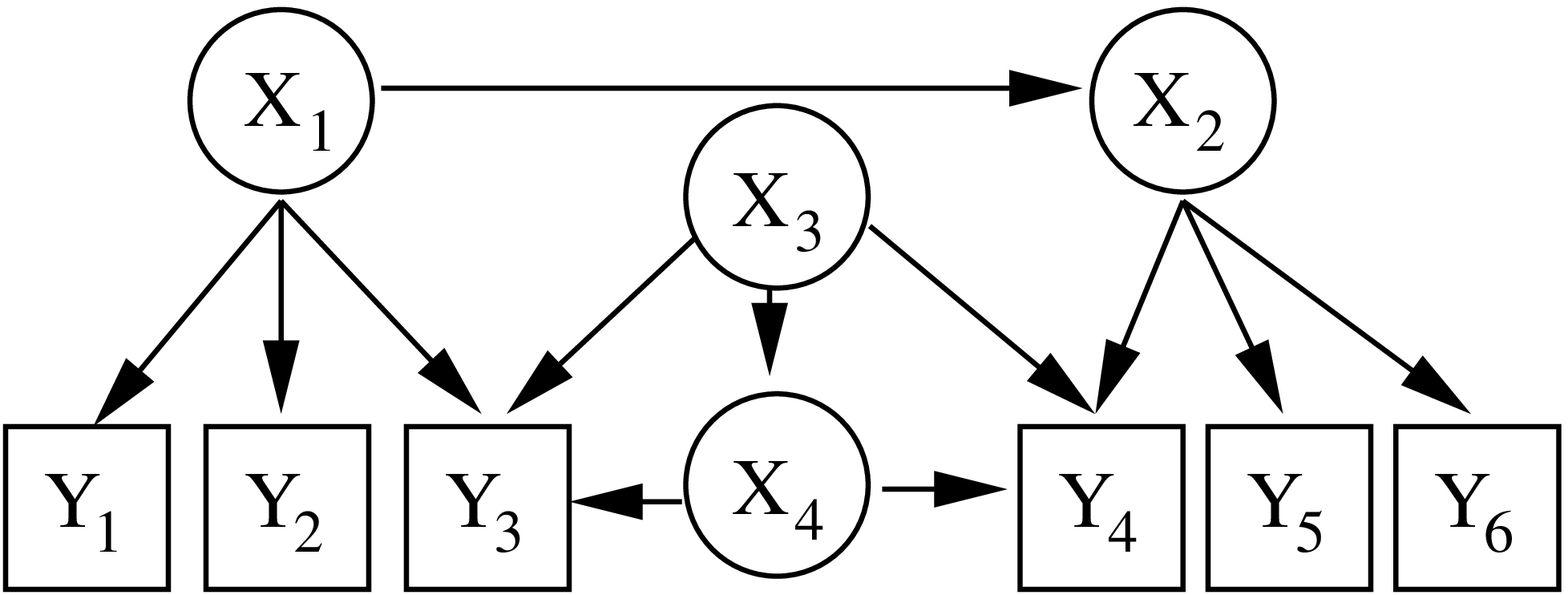,width=5.3cm}\hspace{0.3in} &
\epsfig{file=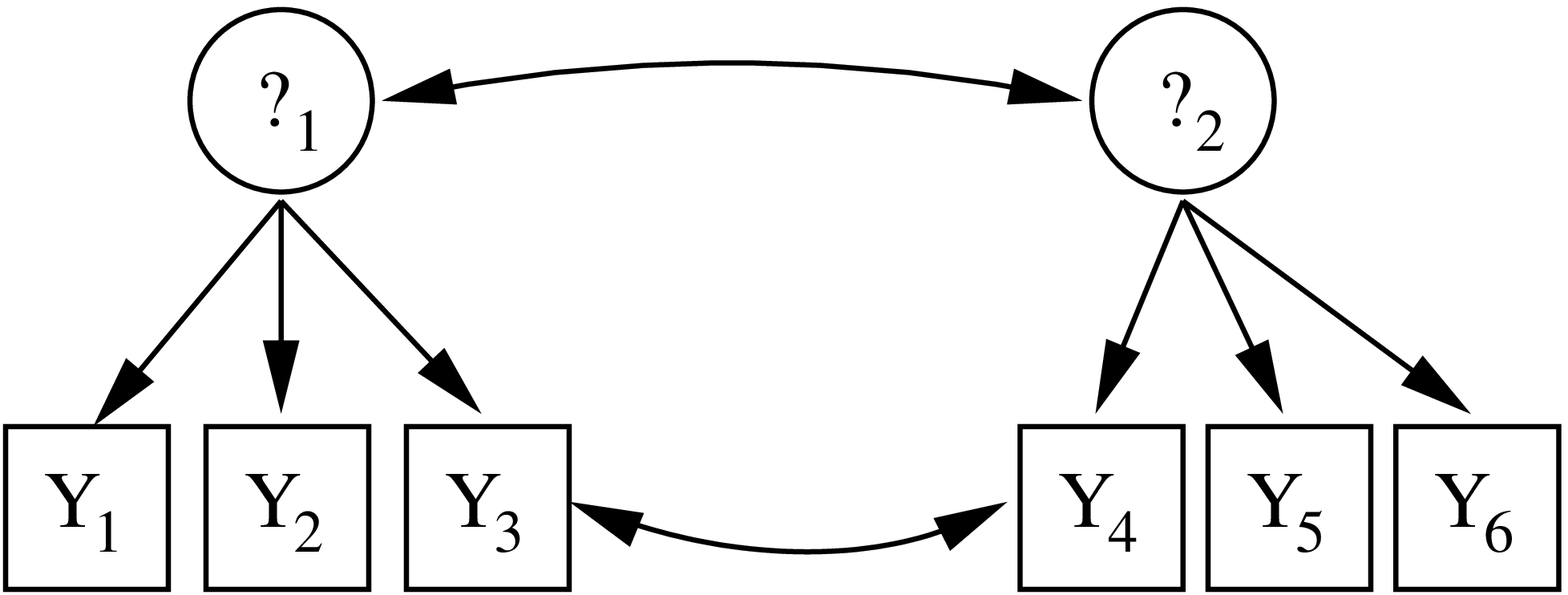,width=5.3cm}\\
(a) & (b)\\
&\\
\epsfig{file=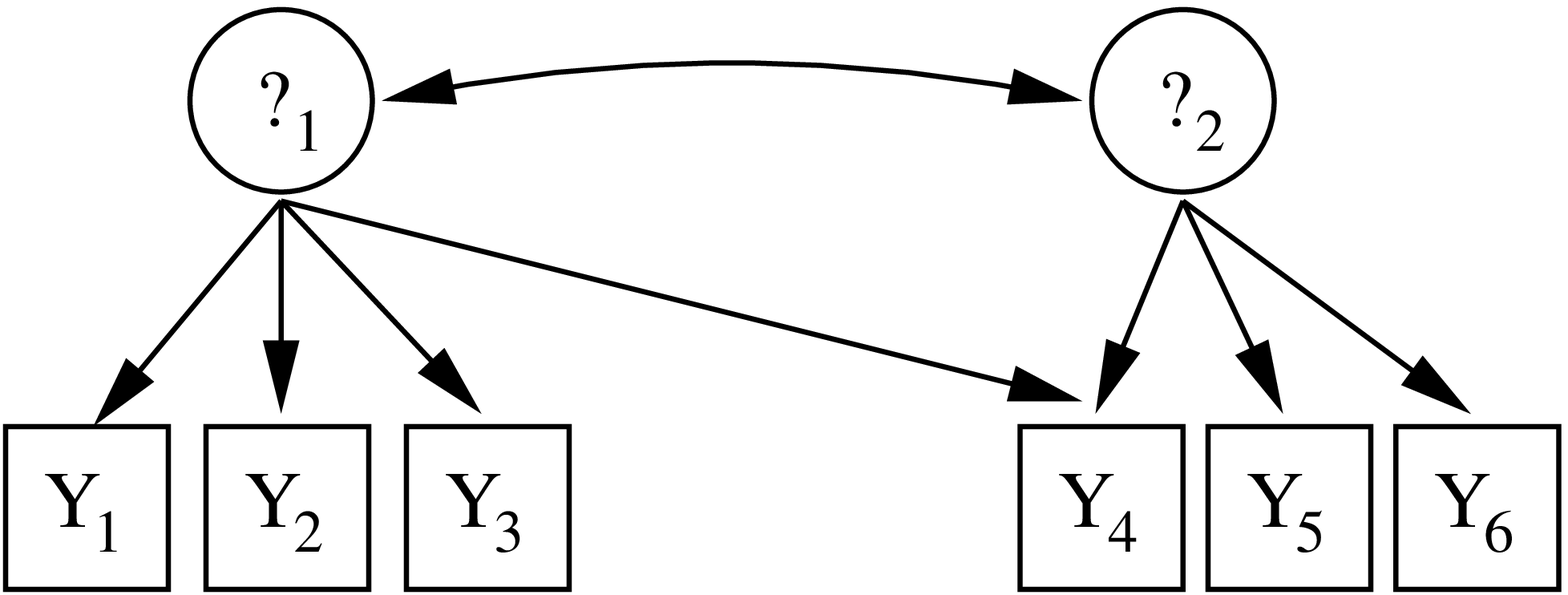,width=5.3cm}\hspace{0.3in} &
\epsfig{file=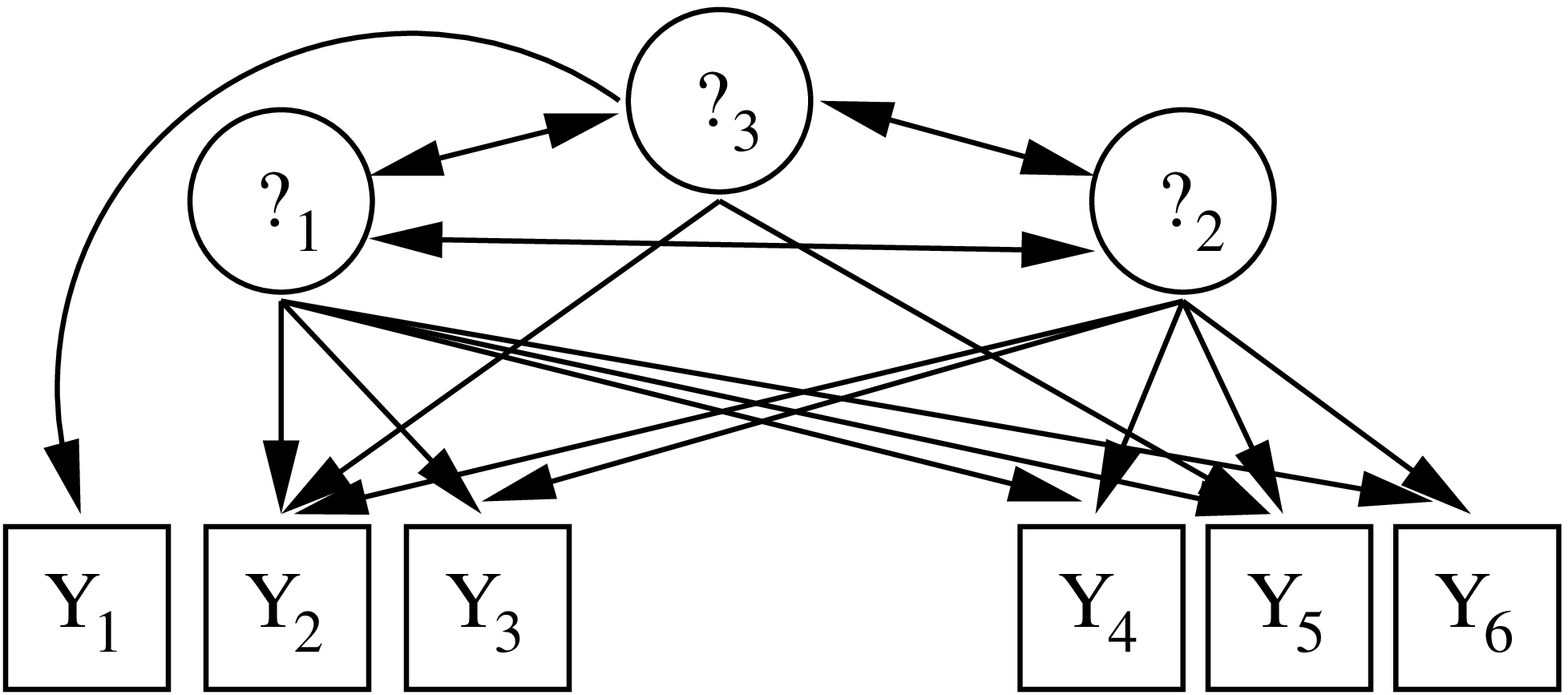,width=5.3cm}\\
(c) & (d)\\
\end{tabular}
\caption{In (a), we show a synthetic
structure from which we generated 200 data points.  Our algorithm is
able to reconstruct the causal graphical structure in (b), which
captures several features of the original model. Bi-directed edges
represent conditional association and the possibility of some
unidentified set of hidden common causes of the corresponding
vertices. Inferred latent variables are not labeled, but can then be
interpreted from the resulting structure and the explicit assumptions
discussed in Section \ref{sec:statement}. In (c), the resulting
structure obtained with exploratory factor analysis using two latent
variables and the promax rotation technique.  In (d), the factor
analysis result we got when attempting to fit three latent
variables.}
\label{fig:simple-example}
\end{figure}

We provide an example in Figure \ref{fig:simple-example}. We generated data
from a linear causal model that follows the causal diagram of Figure
\ref{fig:simple-example}(a)\footnote{Coefficients were generated uniformly at
random on the inverval $[-1.5, -0.5] \cup [0.5, 1.5]$ while variances of error
terms were generated uniformly in $[0, 0.5]$.}.
Given data for the observed variables $Y_1, \dots, Y_6$, we ideally
would like to get a structure such as the one in
\ref{fig:simple-example}(b), where the question marks emphasize that labels for the
latent variables should be provided by background knowledge. Notice
that in this contribution our aim is not to learn the structure
connecting the latent variables, and the bi-directed edge in this case
denotes an arbitrary causal connection.

Factor analysis fails to provide sensible answers. Figure
\ref{fig:simple-example}(c) shows a common outcome when we indicate that the
model should have two hidden common causes. There exists no theory
that provides a clear interpretation for these edges. Even worse,
results can easily become meaningless. In Figure
\ref{fig:simple-example}(d), we depict the result of exactly the same
procedure, but where know we allow for three hidden common causes.
The method we describe in our contribution is able to recover
Figure \ref{fig:simple-example}(b).

\section{Problem statement and assumptions}
\label{sec:statement}

Assume that that our data follows a distribution $\mathcal P$
generated according to a directed acyclic causal graph (DAG) $\mathcal
G$ \citep{sgs:00,pearl:00} with observed nodes $\mathbf Y$ and hidden
nodes $\mathbf X$. We also assume that the resulting distribution
$\mathcal P$ is {\it faithful} to $\mathcal G$ \citep{sgs:00}, that is,
a conditional independence constraint holds in $\mathcal P$ if and only
if it holds in $\mathcal G$ (using the common criterion of {\it d-separation} $-$
please refer to Pearl (2000) for a definition and examples). 
These are all standard assumptions from the causal discovery
literature.

Our more particular assumptions are

\begin{itemize}
\item no observed node $Y \in \mathbf Y$ is a parent in $\mathcal G$ of any hidden node
$X \in \mathbf X$;
\item each random variable in $\mathbf Y \cup \mathbf X$ is a linear 
combination of its parents, plus additive
noise, as in Equation (\ref{eq:linear}).
\end{itemize}

The first assumption is motivated by applications in structural
equation modeling \citep{bol:89}, where prior knowledge is used to
distinguish between standard indicators and ``causal indicators,''
which are causes of the latent variables of interest. Both of these
assumptions can be relaxed to some extent, although any claims
concerning the resulting causal structures learned from data will be
weaker. \cite{sil:06} discuss the details.

For the purposes of simplifying the presentation of this chapter, we also introduce
the following two assumptions:

\begin{itemize}
\item no observed node $Y \in \mathbf Y$ is an ancestor in $\mathcal G$ of any other
observed node;
\item every pair of observed nodes in $\mathbf Y$ has a common latent ancestor\footnote{As a reminder,
this is not the same as having parents in common.} in $\mathcal G$;
\end{itemize}

These two assumptions can be dropped without any loss of generality
\citep{sil:06}, but they will be useful for presentation
purposes. Notice that the latter assumption implies that there are no
conditional independence constraints in the marginal distribution of
$\mathbf Y$. As such, standard causal inference algorithm such as the
PC algorithm \citep{sgs:00} cannot provide any information. 

Notice also that we do not assume any other form of background knowledge
concerning the number of latent variables or particular information
concerning which observed variables have common hidden parents.

Having clarified all assumptions on which our methods rely, the
problem we want to solve can be formalized. Let the {\it measurement
model} $\mathcal M$ of $\mathcal G$ be a graph given by all vertices
of $\mathcal G$, and the edges of $\mathcal G$ that connect
latent variables to observed variables. In order to be agnostic with
respect to the causal structure among latent variables, we connect
each pair of latent variables by a bi-directed edge as a general
symmetric representation of dependency. Ideally, given the distribution
$\mathcal P$ over the observed variables and that our assumptions
hold, we would like to reconstruct $\mathcal M$. Since $\mathcal P$
has to be estimated from the data, it is of practical interest to use
only features of $\mathcal P$ that can be easily estimated while also
accounting for Gaussian distributions. As such, we rephrase our
problem as learning $\mathcal M$ given $\Sigma$, the covariance matrix
of $\mathbf Y$.

However, in general this is only possible if the true model entails
that $\Sigma$ is constrained in ways that cannot be explained by other
models. For instance, if there are more latent variables than observed
variables, and each latent variable is a parent of all elements of
$\mathbf Y$, then $\Sigma$ has no constraints and an infinite number of 
models will be compatible with the data.

\cite{sil:06} formalize the problem by extracting only 
{\it pure measurement submodels} of the true model, subgraphs of
$\mathcal M$ where each observed variable $Y$ has a single parent, and
where this parent d-separates $Y$ from all other vertices of the submodel in
$\mathcal G$. Such single-parent vertices are also called {\it pure
indicators}. Moreover, the output of the procedure described by
\cite{sil:06} only generates submodels where each latent variable has at
least three pure indicators. If such models exist, they can be
discovered given $\Sigma$. The scientific motivation is that many
datasets studied through structural model analysis and factor
analysis support the existence of pure measurement submodels. As we
mentioned in the previous section, methods for providing ``simple
structures'' in factor analysis are hard to justify unless some pure
measurement submodel exists. Therefore, it would be hard to justify factor
analysis as a more flexible approach, since its output would be
unreliable anyway. An important advantage of the causal discovery
approach discussed here is that it knows its limitations.

Our contribution is to extend the work of Silva {\it et al.} by
allowing several ``impurities'' in the output of our new procedure. To
give an example where this is necessary, consider Figure
\ref{fig:simple-example}(a) again. It is not possible to include both
latent variables using the procedure of Silva et {\it et al.}: if
latent variables $X_1$ and $X_2$, and their respective three
indicators, are included, it turns out $Y_3$ is not d-separated from
$Y_4$ by neither $X_1$ nor $X_2$.  The best \cite{sil:06} can do
is to include, say, $X_1$ and its indicators, plus one of its
descendants as an indirect indicator which does not violate the
separations in the true model. For instance, the model with edges $X_1
\rightarrow Y_i$, for $i \in \{1, 2, 3, 5\}$ and no other variable, satisfies
this condition. In contrast, the new procedure described here is able
to generate Figure \ref{fig:simple-example}(b).

In practice, $\Sigma$ has to be estimated from data. In the discussion
that follows, we assume that we know $\Sigma$ so that we can
concentrate on the theory and the main ideas. Section
\ref{sec:experiments} provides methods to deal with an
estimate of $\Sigma$.

\subsection{Description of output}

Our output is a {\it measurement pattern} $\mathcal M_P$ which, under
the above specified assumptions and given the population matrix
$\Sigma$ of a set of observed variables $\mathbf Y$, provides provably
correct causal claims concerning the true structure $\mathcal
M$. The measurement pattern is a directed mixed graph with labeled
edges (as explained below), with hidden nodes $\{L_i\}$ and observed
nodes that form a subset of $\mathbf Y$. $\mathcal M_P$ includes
directed edges from latent variables to observed variables, and
bi-directed edges between observed variables. 

Before introducing the new procedure in Section \ref{sec:procedure},
we formalize the causal claims that a measurement pattern
$\mathcal M_P$ provides:

\begin{enumerate}
\item each hidden variable $L_i$ in $\mathcal M_P$ corresponds to some hidden variable
$X_j$ in $\mathcal G$. In the items below, we call this variable $X(L_i)$;
\item if $Y_i$ is not a child of latent variable $L_j$ in $\mathcal M_P$, then
$Y_i$ is independent of $X(L_j)$ in $\mathcal G$ given its parents in $\mathcal M_P$;
\item given {\it any} pure measurement submodel of $\mathcal M_P$ with at least three 
indicators per latent variable, and a total of at least four observed variables, 
then {\it at most one} of the latent-to-indicator edges $L_i \rightarrow Y_j$ 
does not correspond to the true causal relationship in $\mathcal G$. That is, it is
possible that for one pair, $X(L_i)$ is not a cause of $Y_j$ and/or the relationship is confounded;
\end{enumerate}

\begin{figure}[t]
\begin{tabular}{cc}
\epsfig{file=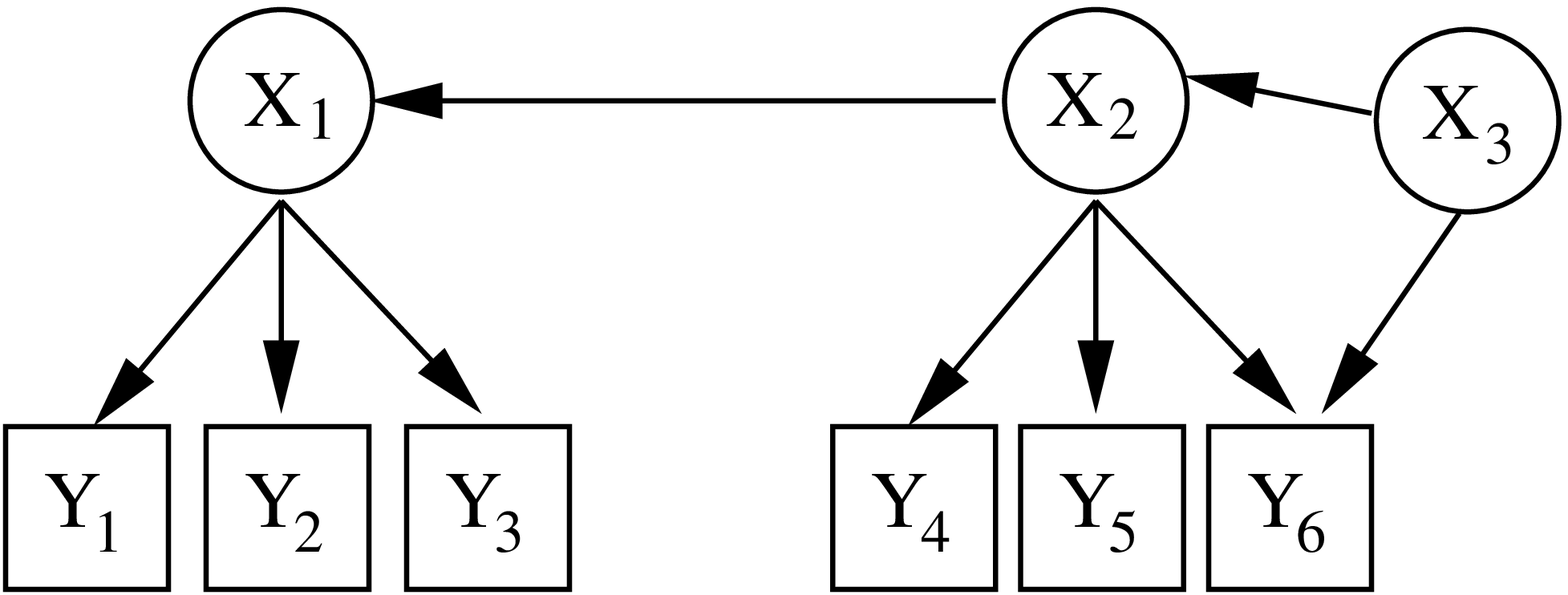,width=5.3cm}\hspace{0.3in} &
\epsfig{file=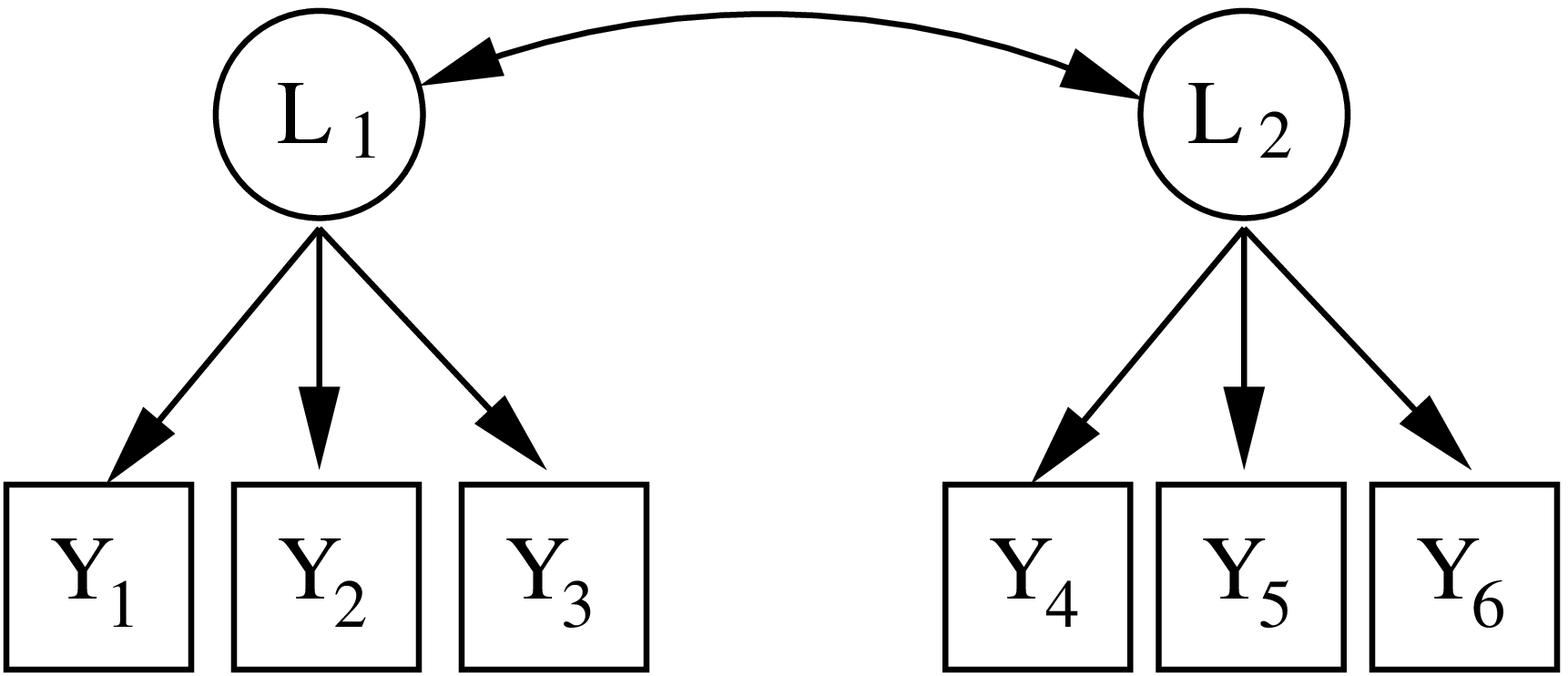,width=5.3cm}\\
(a) & (b)\\
\end{tabular}
\caption{A covariance matrix obtained from the true model in (a)
will result in the structure shown in (b). This structure should
be interpreted as making claims about an equivalence class of models:
in this case, at most one of the directed edges does not correspond to the
exact claim that there is an unconfounded causal relationship between the
latent variable and its respective child. But notice that there are only
7 models compatible with this claim, instead of the $2^6=32$ possibilities
(each of the six relations $L_i \rightarrow Y_j$ being confounded or not)
that the same adjacencies could provide.}
\label{fig:confounder}
\end{figure}

The last item needs to be clarified with an example, since it is not
intuitive. Let Figure \ref{fig:confounder}(a) be a true causal structure
from which we can measure the covariance matrix of $Y_1, \dots,
Y_6$. The structure reported by our procedure is the one in Figure
\ref{fig:confounder}(b). 5 out of 6 edges correspond to the correct
causal statement, except $L_2 \rightarrow Y_6$ (which should be
confounded). We cannot know which one, but at least we know this is
the case. As in any causal discovery algorithm
\citep{sgs:00,pearl:00}, background knowledge is necessary to
refine the information given by an equivalence class of graphical
structures.

Finally, edges are labeled as ``confirmed'' (they do correspond to
actual paths in the true graph) or ``unconfirmed'' (we cannot decide
whether a corresponding path exists in the true graph). In the next
section we clarify how ``unconfirmed'' edges appear. Unless otherwise
stated, all other edges are ``confirmed'' edges.

\section{An algorithm for inferring an impure measurement model}
\label{sec:procedure}

Let a {\it one-factor submodel} of $\mathcal G$ be a set composed of one hidden
variable $X$ and four observed variables $\{Y_A, Y_B, Y_C, Y_D\}$ such that
$X$ d-separates all four observations in $\mathcal G$. 

One-factor submodels play an important role in our procedure.  A
vertex $Y_i$ will be included in our output measurement pattern
$\mathcal M_P$ if and only if it belongs to some one-factor submodel
of $\mathcal G$. Also, $X$ will correspond to some output latent if
and only if it belongs to some one-factor submodel. Figure
\ref{fig:simple-example}(a) illustrates the concept: the sets $\{X_1,
Y_1, Y_2, Y_3, Y_5\}$ and $\{X_2, Y_4, Y_5, Y_6, Y_1\}$ are one-factor
submodels. No one-factor submodels exist for $X_3$ and $X_4$.

This fact should not be surprising. It is well-known in the structural
equation modeling literature that the folllowing model is testable:

\begin{equation}
\label{eq:onefactor}
Y_i = \lambda_iX + \epsilon_i
\end{equation}

\noindent where $i \in \{1, 2, 3, 4\}$, and $X$ and $\{\epsilon_i\}$ are mutually
independent Gaussian variables of zero mean. This corresponds to a Gaussian causal
network with corresponding edges $X \rightarrow Y_i$. Adding an extra
edge, and hence a new parameter, would remove one degree of freedom
and make the model undistinguishable from models with two latent
variables \citep{sil:06}.

One way to characterize which constraints are entailed by this model is
by writing down the {\it tetrad constraints} of this structure. If $\sigma_{ij}$
is the covariance of $Y_i$ and $Y_j$, and $\sigma^2_X$ is the variance of $X$,
then for the model (\ref{eq:onefactor}) the following identify holds:

\begin{equation}
\sigma_{12}\sigma_{34} = \lambda_1\lambda_2\sigma^2_X \times \lambda_3\lambda_4\sigma^2_X =
\lambda_1\lambda_3\sigma^2_X \times \lambda_2\lambda_4\sigma^2_X = \sigma_{13}\sigma_{34}
\end{equation}

Similarly, $\sigma_{12}\sigma_{34} = \sigma_{14}\sigma_{23}$. For a
set of four variables $\{Y_A, Y_B, Y_C, Y_D\}$, we represent the
statement $\sigma_{AB}\sigma_{CD} = \sigma_{AC}\sigma_{BD} =
\sigma_{AD}\sigma_{BC}$ by the predicate $\mathcal T(ABCD)$.
Notice this is entailed by the graphical structure, since the
relationship does not depend on the precise values of $\{\lambda_i\}$
or $\sigma^2_X$.

For the causal discovery goal, however, the relevant concept is the converse:
given observable constraints that can be tested, which causal structures are
compatible with them? Concerning one-factor submodels, the converse has been
proved\footnote{To avoid unnecessary repetition, from now on we establish the
convention that all results use the assumptions of Section \ref{sec:statement},
without explicitly mentioning them in the theoretical development.} 
by Silva {\it et al.} (2006):\\

\noindent{\bf Fact 1}\hspace{0.1in} If $\mathcal T(ABCD)$ is true, then there is
a latent variable in $\mathcal G$ that d-separates $\{Y_A, Y_B, Y_C, Y_D\}$.\\

For example, in Figure \ref{fig:confounder}(a), $X_1$ d-separates
$\{Y_1, Y_2, Y_3, Y_4\}$, although it is not a cause of $Y_4$.  A
result such as Fact 1 is important for discovering latent variables,
but it is of limited use unless there are ways of ruling out the
possibility that some latent variables are causes of some indicators.
It turns out that the $\mathcal T(\cdot)$ constraint
can also be used for this purpose. 

Consider Figure \ref{fig:confounder}(a) again. If we pick all three indicators
of one latent variables along with some indicator of the other latent variables,
we have a one-factor model that passes the conditions of Fact 1. One possibility
is that all six indicators are pure indicators of a single latent cause: after all,
each pair $\{Y_A, Y_B\}$ is d-separated by some single latent variable.
However, this does not tell us whether the latent variable that separates one group
is the same as the one that separates another group. This is clear
from Figure \ref{fig:confounder}(a): $X_1$ d-separates any pair in $\{Y_1, Y_2, Y_3\}
\times \{Y_4, Y_5, Y_6\}$. However, it does not d-separate any pair in 
$\{Y_4, Y_5, Y_6\} \times \{Y_4, Y_5, Y_6\}$. We have to deduce this
information without looking at the true graph, but only at the marginal
covariance matrix of $\mathbf Y$.

One way of discarding connections from latents to indicators, and
deducing that two unobserved variables are not the same, is given by
the following result:\\

\noindent{\bf Fact 2}\hspace{0.1in} Consider the observed variables $\{Y_A, Y_B, Y_C, Y_D, Y_E, Y_F\}$.
If both \\$\mathcal T(ABCD)$ and $T(ADEF)$ are true, but 
$\sigma_{AB}\sigma_{DE} \ne \sigma_{AD}\sigma_{BE}$, then $Y_A$ and $Y_D$ cannot have
any common parent in $\mathcal G$.\\

A detailed proof is given by Silva {\it et al.} (2006). The intuitive
explanation is that, if $Y_A$ and $Y_D$ did have a common parent (say,
$X_{AD}$), then this latent variable would be precisely the one,
and only one, responsible by both constraints $\mathcal T(ABCD)$ and $\mathcal
T(ADEF)$. It would not be hard to show that this would imply
$\sigma_{AB}\sigma_{DE} = \sigma_{AD}\sigma_{BE}$, contrary to the
assumption.

Notice that these two results are already enough to find a pure
measurement submodel. The general skeleton of the procedure is to
find a partition $\{\mathbf M_1, \dots, \mathbf M_C\}$ such that
$\cup_{i = 1}^C \mathbf M_i \subseteq \mathbf Y$ and

\begin{enumerate}
\item elements in $\mathbf M_i$ are d-separated by some hidden variable (using Fact 1);
\item elements in $\mathbf M_i$ and $\mathbf M_j$ cannot have common parents (using Fact 2).
\end{enumerate}

Many more details need to be figured out in order to build an
equivalence class of pure measurement models with three indicators per
latent variable, but this is the general idea. What is missing from
this procedure is a way of coping with impure measurement models so
that a structure such as the one in Figure \ref{fig:simple-example}(b)
can be obtained. We now introduce the first theoretical results that
accomplish that.

\subsection{Finding impure indicators}

Consider what can happen if we observe the covariance matrix generated
by the model of Figure \ref{fig:simple-example}(a). We know that there
is no single latent variable that d-separates (say) $\{Y_1, Y_2, Y_3,
Y_4\}$. However, we know that there is some hidden $L$ that d-separates
$\{Y_1, Y_2, Y_3, Y_5\}$, as well as some hidden $L'$ that
d-separates $\{Y_1, Y_4, Y_5, Y_6\}$. So far, it could be the result
of a graph such as the following:

\begin{center}
\epsfig{file=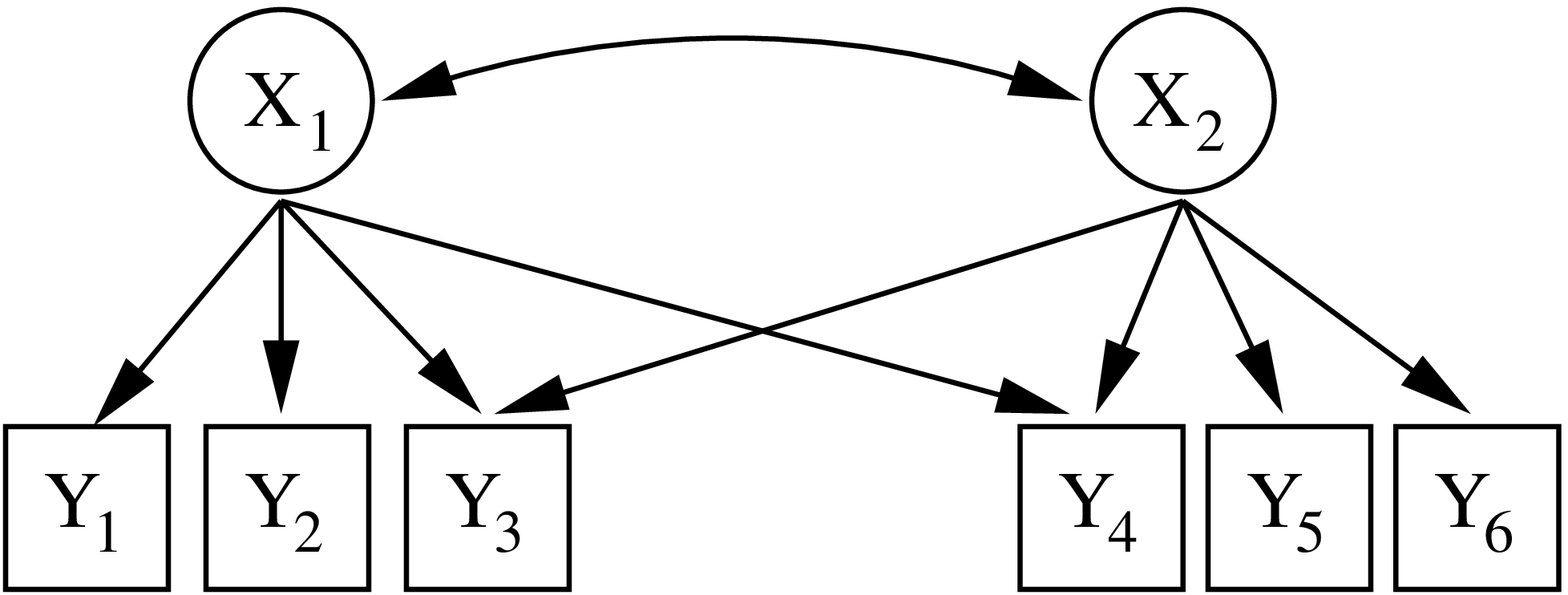,width=5.3cm}
\end{center}

However, we cannot stop here and report this as a possible solution:
we will get an inconsistent estimate for the covariance of the latent
variables, which can lead to wrong conclusions about the causal
structure of the latents. We would like to account for the possibility
that the impurities arise not from our identified latents, but from
some other source. This is the result summarized by Lemma 3:\\

\noindent{\bf Lemma 3}\hspace{0.1in} 
{\it Consider the observed variables $\{Y_A, Y_B, Y_C, Y_D, Y_E, Y_F\}$.
If the folllowing predicates are true:

\begin{center}
$\mathcal T(ABCE)$, $\mathcal T(ABCF)$, $\mathcal T(ADEF)$, $\mathcal T(BDEF)$
\end{center}

and the following predicates are false

\begin{center}
$\mathcal T(ABEF)$, $\mathcal T(ABCD)$, $\mathcal T(CDEF)$
\end{center}

then in the corresponding causal graph $\mathcal G$, we have that:

\begin{itemize}
\item $\mathcal G$ contains at least two different latent variables, $L_1$ and $L_2$;
\item $L_1$ d-separates all pairs in $\{Y_A, Y_B, Y_C\} \times \{Y_D, Y_E, Y_F, L_2\}$,
      except $Y_C \times Y_D$;
\item $L_2$ d-separates all pairs in $\{Y_A, Y_B, Y_C, L_1\} \times \{Y_D, Y_E, Y_F\}$,
      except $Y_C \times Y_D$;
\item $Y_C$ and $Y_D$ have extra hidden common causes not in $\{L_1, L_2\}$.
\end{itemize}
}

A formal proof of a slightly more general result is given by
\cite{silva:06d}. The core argument is as follows. The existence of
$L_1$ and $L_2$ follows from Fact 1 and the constraints $\mathcal
T(ABCE)$ and $\mathcal T(ADEF)$. That $L_1 \ne L_2$ follows from Fact
2 and the fact that $\mathcal T(ABEF)$ is false. The other
d-separations follow from Fact 1 and the given tetrad
constraints. Finally, if $Y_C$ and $Y_D$ did not have any other hidden
common cause, we could not have both $\mathcal T(ABCD)$ and $\mathcal
T(CDEF)$ falsified at the same time, contrary to our hypothesis.

\begin{figure}[t]
\begin{tabular}{cc}
\epsfig{file=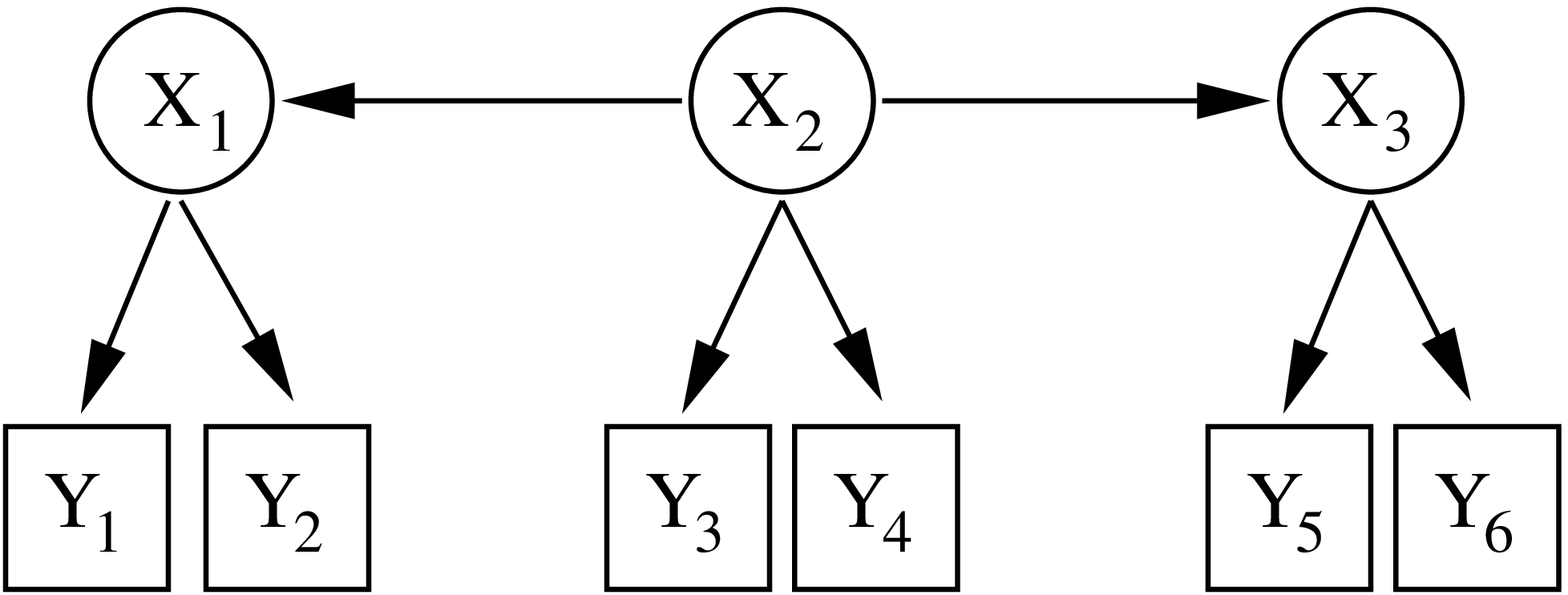,width=5.3cm}\hspace{0.3in} &
\epsfig{file=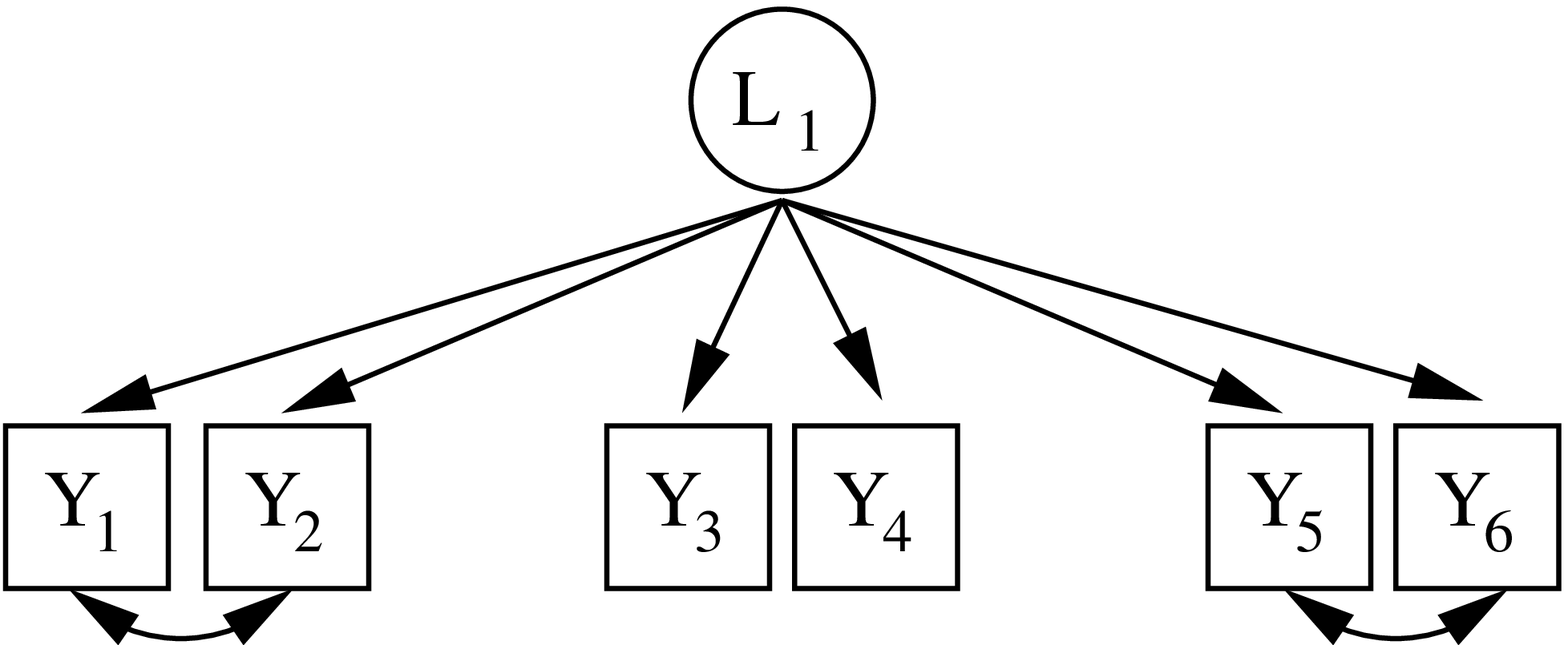,width=5.3cm}\\
(a) & (b)\\
\end{tabular}
\caption{A model such as the one in (a) generates the measurement pattern in (b).
Notice that the indication of extra hidden common causes, as represented by, e.g.,
$Y_1 \leftrightarrow Y_2$, does not imply that these unrepresented causes are
independent of the represented ones. Notice also that if the edge $X_1 \leftarrow X_2$
was switched in (a) to $X_1 \rightarrow X_2$, the corresponding pattern would still
be the one in (b). In this case, it is clear that both edges $L_1 \rightarrow Y_1$ and
$L_1 \rightarrow Y_2$ are not representing the actual causal directions of the
true graph. However, the corresponding measurement pattern claim is about pure submodels.
In this case, a pure submodel could be the one containing $Y_1, Y_3, Y_4$ and $Y_5$
only. One edge, $L_1 \rightarrow Y_1$, still does not explicitly indicate the confounding
given by $X_1$, but this is compatible with the measurement pattern description.}
\label{fig:unrepresented}
\end{figure}

Notice that we never claim that the implicit latent variables represented
by bi-directed edges are independent of the discovered latent variables.
Figure \ref{fig:unrepresented} illustrates a case.

The second type of impurity we will account for nodes that
have more than one represented latent parent. \\

\noindent{\bf Lemma 4}\hspace{0.1in} 
{\it Consider the observed variables $\{Y_A, Y_B, Y_C, Y_D, Y_E, Y_F, Y_G\}$.
If the folllowing predicates are true:

\begin{center}
$\mathcal T(ABCK)$, for $K \in \{D, E, F, G\}$;
$\mathcal T(KEFG)$, for $K \in \{A, B, C, D\}$;
\end{center}

and the following predicates are false

\begin{center}
$\mathcal T(K_1K_2K_3K_4)$, for $\{K_1, K_2\} \subset \{A, B, C\}$, 
$\{K_3, K_4\} \subset \{E, F, G\}$;

$\mathcal T(ADEF)$, $\mathcal T(ABDE)$
\end{center}

then in the corresponding causal graph $\mathcal G$, we have that:

\begin{itemize}
\item $\mathcal G$ contains at least two different latent variables, $L_1$ and $L_2$;
\item $L_1$ d-separates all pairs in $\{Y_A, Y_B, Y_C, Y_D\}$;
\item $L_2$ d-separates all pairs in $\{Y_D, Y_E, Y_F, Y_G\}$;
\item $L_1$ d-separates all pairs in $\{Y_A, Y_B, Y_C\} \times \{Y_E, Y_F, Y_G, L_2\}$, 
      but not $Y_D \times L_2$;
\item $L_2$ d-separates all pairs in $\{Y_A, Y_B, Y_C, L_1\} \times \{Y_E, Y_F, Y_G\}$,
      but not $Y_D \times L_1$;
\end{itemize}
}

The nature of this result complements the previous one: instead of
searching for evidence to remove edges from latents into indicators,
this result provides identification of edges that {\it cannot} be removed.

\begin{figure}[t]
\begin{tabular}{cc}
\epsfig{file=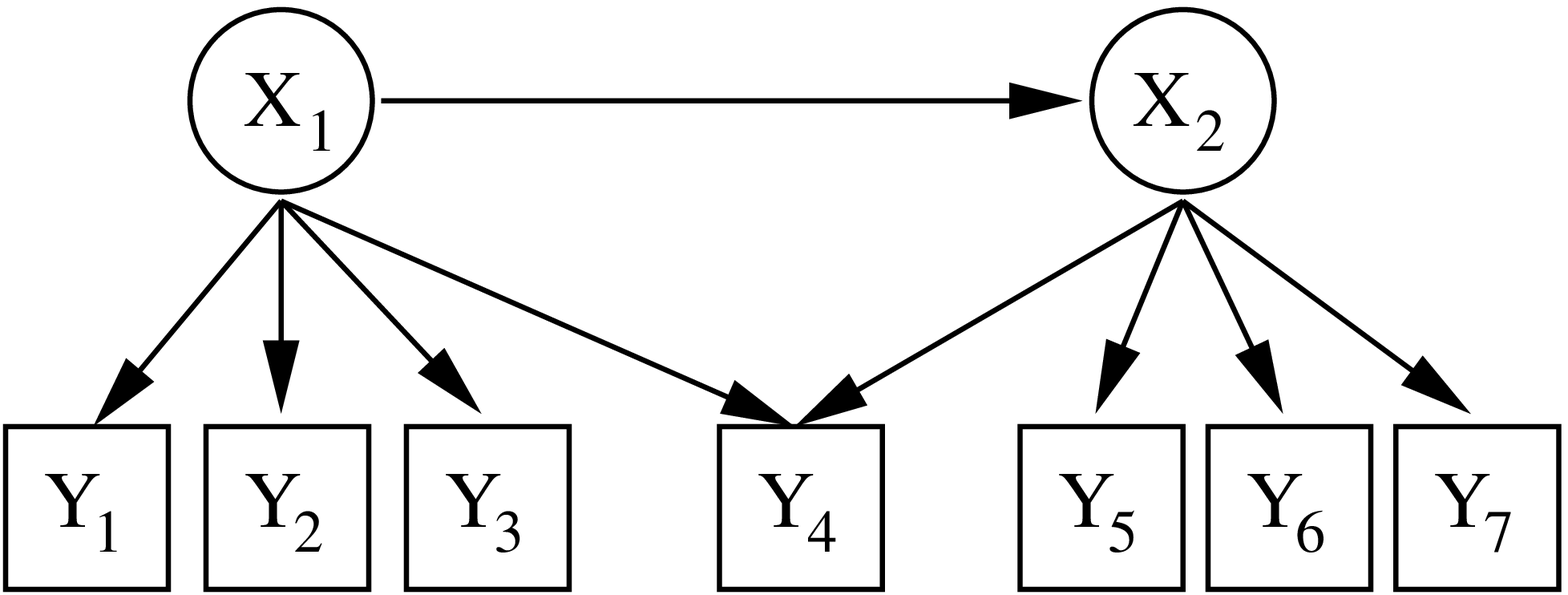,width=5.3cm}\hspace{0.3in} &
\epsfig{file=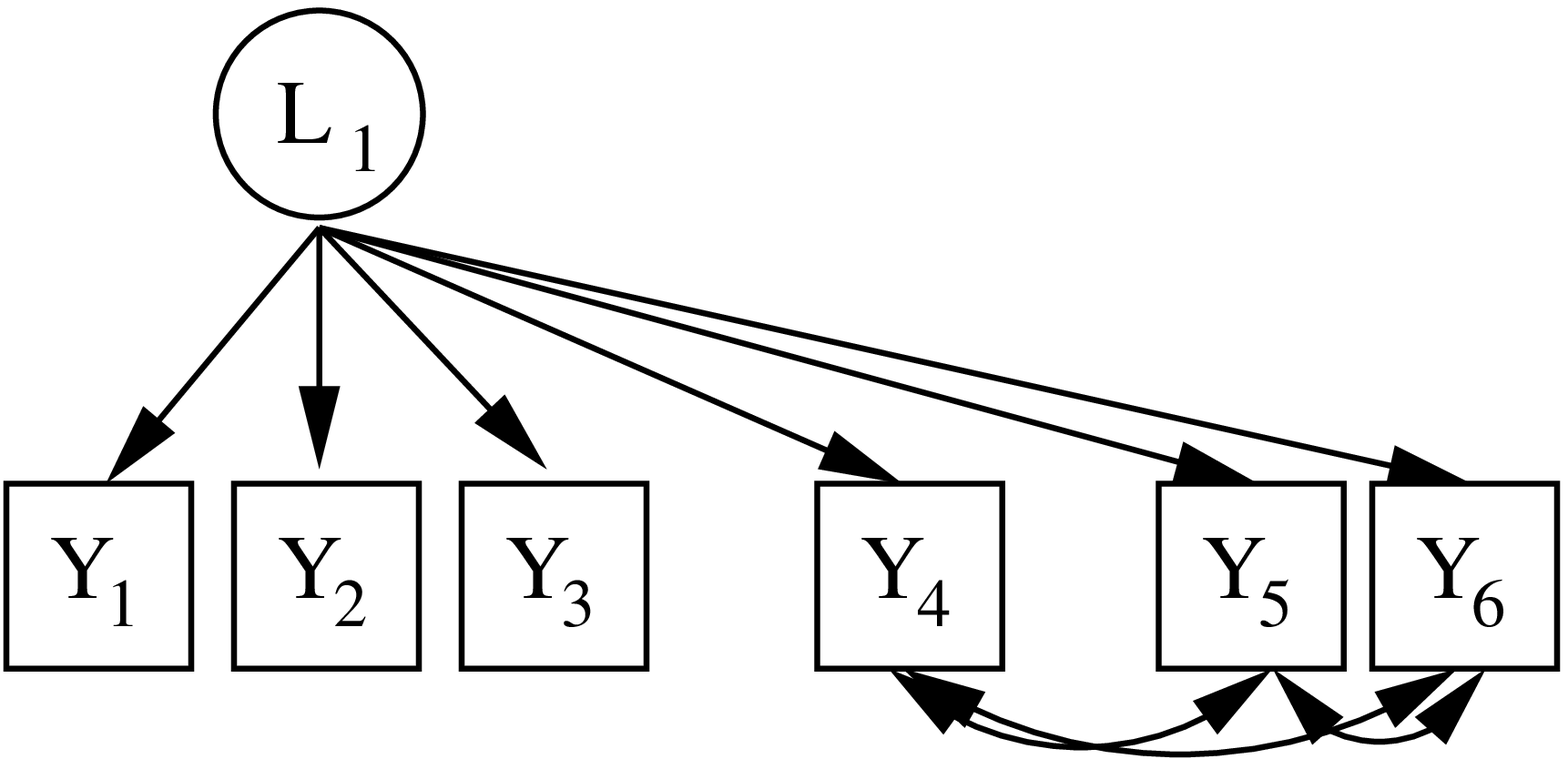,width=5.3cm}\\
(a) & (b)\\
\end{tabular}
\caption{The graph in (a) can be rebuild exactly. However, without a fourth indicator of
$X_2$ (e.g., $Y_7$), this latent variable can not be detected and the result would be the
graph in (b).}
\label{fig:multi}
\end{figure}

The argument again exploits Facts 1 and 2. A more detailed proof is
given by \cite{silva:06d}. Notice the need for extra indicators in
this case: this is another illustration of the need for one-factor
models for each latent variable. Without $Y_7$ in the example of
Figure \ref{fig:multi}(a), the result would be the measurement pattern
of Figure \ref{fig:multi}(b).

\begin{figure}[t]
\begin{tabular}{cc}
\epsfig{file=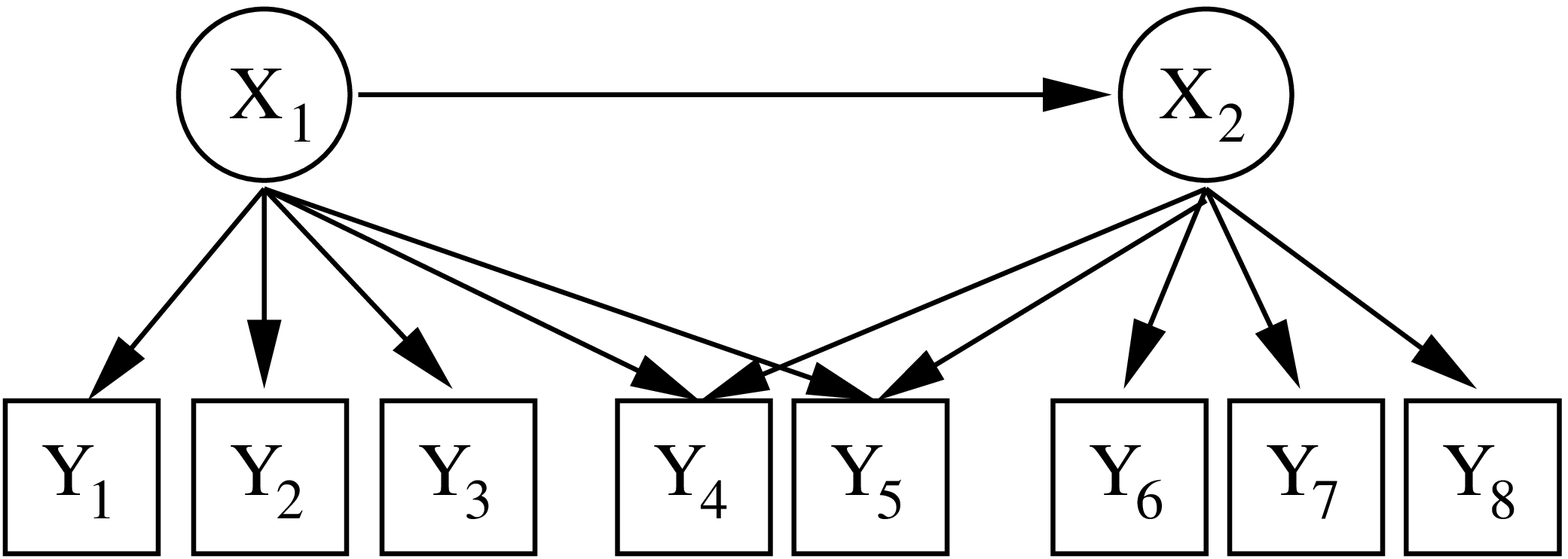,width=5.3cm}\hspace{0.3in} &
\epsfig{file=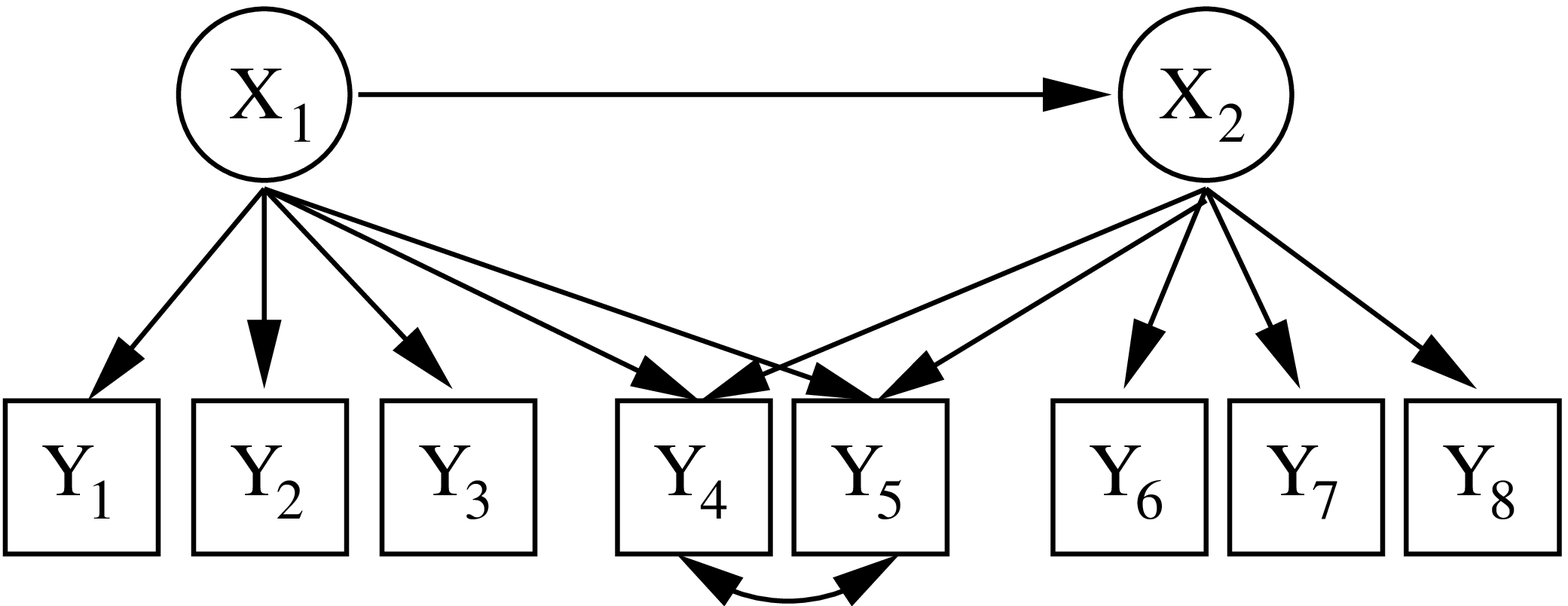,width=5.3cm}\\
(a) & (b)\\
\end{tabular}
\caption{The graph in (a) is not fully identifiable with tetrad constraints only.
A conservative measurement pattern needs to report the structure in (b).}
\label{fig:multi2}
\end{figure}

Notice that if there are indicators that share more than one common
parent in $\mathcal G$, we cannot separate them (i.e., avoid a
bi-directed edge) even if their parents are identified in the model
using tetrad constraints only. Figure \ref{fig:multi2} illustrates
what the measurement pattern should report. Using higher-order
constraints than tetrad constraints might be of help in this situation
\citep{sullivant:08}, but this is out of the scope of the current contribution.

To summarize:

\begin{itemize}
\item Fact 1 provides the evidence to include latent variables;
\item Fact 2 provides the evidence to distinguish between different latent variables;
\item Lemma 3 allows the removal of extra edges from latents into indicators and proves
      the necessity of some bi-directed edges;
\item Lemma 4 proves the necessity of some edges from latents into indicators, but does
      not prove the necessity of adding some bi-directed edges;
\end{itemize}

\subsection{Putting the pieces together}

So far, we have described how to identify particular pieces of
information about the underlying causal graph.  While those results
allow us to identify isolated latent variables and to remove or
confirm particular connections, we need to combine such pieces within
a measurement pattern. Unlike the procedure of \cite{sil:06},
this pattern should be able to represent several pure measurement submodels
within a single graphical object and to possibly include more latent variables
than any pure model.

In this section, we assume that we have the population covariance
matrix $\Sigma$. We start by finding groups of variables that are
potential indicators of a single latent variable. We first build an
auxiliary undirected graph $\mathcal H$ as follows:\\

\noindent{\sc InitialPass}: this procedure returns an undirected graph $\mathcal H$.

\begin{enumerate}
\item let $\mathcal H$ be a fully connected undirected graph with nodes $\mathbf Y$;
\item for all groups of six variables $\{Y_A, Y_B, Y_C, Y_D, Y_E, Y_F\}$ that form a 
      clique in $\mathcal H$, if $\mathcal T(ABCD)$ and $\mathcal T(ADEF)$ are true
      but $\sigma_{AB}\sigma_{DE} \ne \sigma_{AD}\sigma_{BE}$, remove the edge 
      $Y_A - Y_D$;
\item if for a given $Y_A$ in $\mathcal H$ there is no triplet $\{Y_B, Y_C, Y_D\}$ 
      such that $\mathcal T(ABCD)$ holds, then remove $Y_A$ from 
      $\mathcal H$, since there will be no one-factor model including $Y_A$;
\item return $\mathcal H$.
\end{enumerate}

Notice that if two vertices are not adjacent in $\mathcal H$, they
cannot possibly be children of the same latent variable (it follows
from Fact 2). This motivates us to look for one-factor models within
cliques of $\mathcal H$ only. In Step 3, we discard variables not in
one-factor models, since nothing informative can be claimed about them
using our methods.

In the next step, we obtain a set of tentative subgraphs, where each
subgraph contains a single latent variable and its indicators:\\

\noindent{\sc SingleLatents}: given $\mathcal H$, 
this procedure returns a set $\mathcal S$ of graphs
 with a single latent variable each.

\begin{enumerate}
\item initialize $\mathcal S$ as the empty set
\item for each clique $\mathcal C$ in $\mathcal H$ 
\item \hspace{0.2in} if there is no $\{Y_A, Y_B, Y_C\} \subset \mathcal C$ and $Y_D \in \mathbf Y$ such
      that $\mathcal T(ABCD)$ 
\item[]\hspace{0.2in} holds, continue to next clique
\item \hspace{0.2in} create a graph $\mathcal G_i$ with latent vertex $L_i$, $i = |\mathcal S| + 1$, and
\item[]\hspace{0.2in} with children given by $\mathcal C$
\item \hspace{0.2in} for each $\{Y_A, Y_B\} \subset \mathcal C$, if there is no
     $Y_C \in \mathcal C$ and $Y_D \in \mathbf Y$ such that
\item[]\hspace{0.2in} $\mathcal T(ABCD)$ is true,
     then add edge $Y_A \leftrightarrow Y_B$ to $\mathcal G_i$.
     Mark this edge as 
\item[]\hspace{0.2in} ``unconfirmed'';  
\item \hspace{0.2in} add the new graph to $\mathcal S$;
\item return $\mathcal S$.
\end{enumerate}

By Fact 1, every single latent variable created in this procedure
exists in $\mathcal G$. The rationale for Step 5 is that $L_i$ does
not d-separate $Y_A$ and $Y_B$. It is possible to confirm many of such
edges using an argument similar to Lemma 4, but we leave out a
detailed analysis to simplify the presentation\footnote{An example: in
Figure \ref{fig:multi}(b), all bi-directed edges can be confirmed,
because each of $\{Y_4, Y_5, Y_6\}$ are separated from $\{Y_1, Y_2,
Y_3\}$ by $L_1$. We can therefore isolate the failure of having a
one-factor model composed of $\{L_1, Y_1, Y_2, Y_4, Y_5\}$ down to the
$Y_4 \leftrightarrow Y_5$ edge.}.

Finally, all single graphs are unified into a coherent measurement pattern:\\

\noindent{\sc FindMeasurementPattern}: returns a measurement pattern $\mathcal M_P$ given $\mathcal S$.

\begin{enumerate}
\item let $\mathcal M_P$ be the union of all graphs in $\mathcal S$, where all latents are
      connected by bi-directed edges
\item for every pair $\{\mathcal S_i, \mathcal S_j\} \subset \mathcal S$ do
\item \hspace{0.2in} consider all triplets $\{Y_A, Y_B, Y_C\} \subset \mathcal S_i \cup \mathcal S_j$ 
      such that $\mathcal T(ABCD)$ holds 
\item[]\hspace{0.2in} for some $Y_D$. If such triplets are also 
      in $\mathcal S_i \cap \mathcal S_j$, set the children of $L_j$ 
\item[]\hspace{0.2in} to be children of $L_i$ and discard $L_j$. Set all $L_i \rightarrow Y_k$
      to be ``unconfirmed'' 
\item[]\hspace{0.2in} if $Y_k$ is not in $\mathcal S_i \cap \mathcal S_j$. Continue to next pair;
\item \hspace{0.2in} for every pair $\{Y_C, Y_D\} \subset \mathcal S_i \cap S_j$, add 
      ``unconfirmed'' edge $\mathcal Y_C \leftrightarrow Y_D$ 
\item[]\hspace{0.2in} to $\mathcal M_P$. If Lemma 3 can be applied to
      $\{Y_C, Y_D\}$ where $\{Y_A, Y_B\} \subset S_i$
\item[]\hspace{0.2in} and $\{Y_E, Y_F\} \subset S_j$, then
      remove edges $L_j \rightarrow Y_C$ and $L_i \rightarrow Y_D$ and
\item[]\hspace{0.2in} mark $\mathcal Y_C \leftrightarrow Y_D$ as ``confirmed'';
\item if $Y_j$ has more than one parent, mark all directed edges $L_i \rightarrow Y_j$ unsupported by Lemma 4 as
      ``unconfirmed'';
\item return $\mathcal \mathcal M_P$.
\end{enumerate}

\begin{figure}[t]
\begin{tabular}{cc}
\epsfig{file=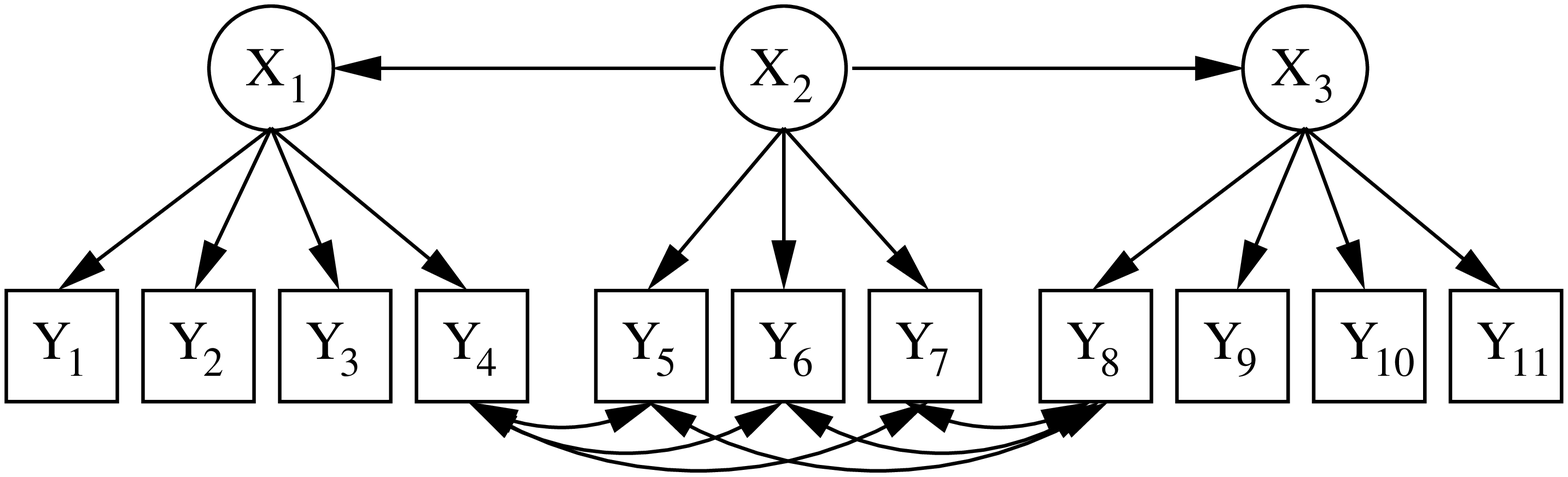,width=5.3cm}\hspace{0.3in} &
\epsfig{file=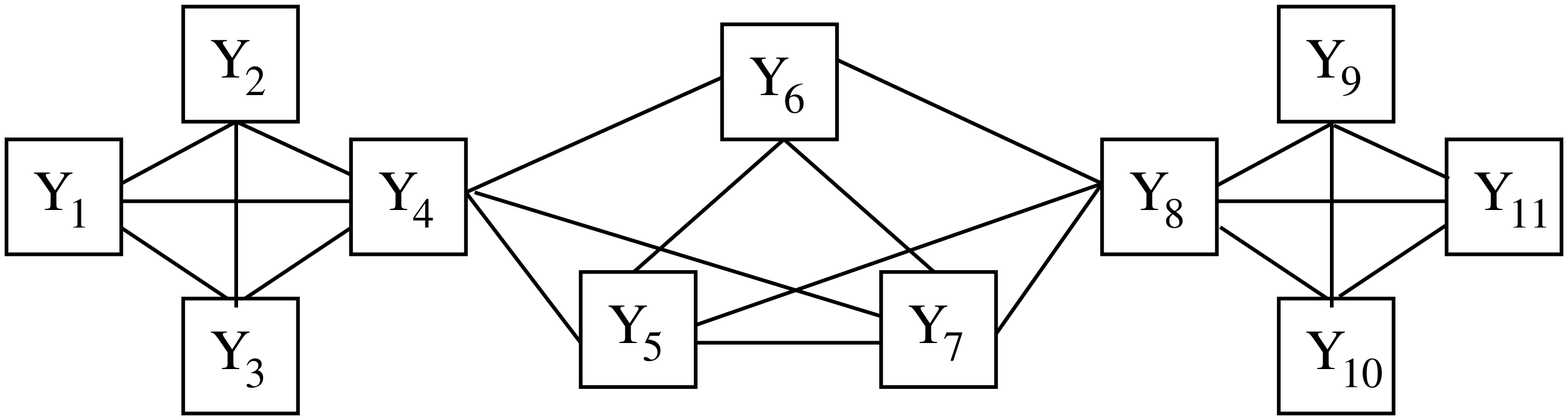,width=5.3cm}\\
(a) & (b)\\
\end{tabular}
\caption{With the true graph being (a), we obtain two cliques of variables
$\{Y_4, Y_5, Y_6, Y_7\}$ and $\{Y_5, Y_6, Y_7, Y_8\}$ in $\mathcal H$
(Figure (b)), since we can
discover that $Y_4$ and $Y_8$ are indicators of different
variables. However, these two cliques are related to the same true
latent $X_2$ and have to be merged. The side-effect is that we cannot
confirm the edges $L_2 \rightarrow Y_4$ and $L_2 \rightarrow Y_8$,
although we know both cannot possibly exist at the same time.}
\label{fig:example-merge}
\end{figure}

The justification for most steps follows directly from our previous
results. To understand Step 3, however, we need an example. In Figure
\ref{fig:example-merge}(a), we have a true model. We can separate
$Y_4$ from $Y_8$ using Fact 2. The result of {\sc FirstPass} is the
graph $\mathcal H$ shown in Figure \ref{fig:example-merge}(b).  Sets
$\{Y_4, Y_5, Y_6, Y_7\}$ and $\{Y_5, Y_6, Y_7, Y_8\}$ are cliques in
$\mathcal H$, but they refer to the same latent variable $X_2$.  There
will be edges $L_2 \rightarrow Y_4$ and $L_2 \rightarrow Y_8$ in the
measurement pattern, but they will not be confirmed edges. Notice that
there might be ways of removing $L_2 \rightarrow Y_4$ and $L_2
\rightarrow Y_8$, but they are out of the scope of our paper.  Our
goal is not to provide complete identification methods, but to show
the main tools and the difficulties of learning impure measurement
models.

\section{Experiments}
\label{sec:experiments}

In this section, we illustrate how the theory can be applied by
analyzing two simple datasets.

In practice, we will not know $\Sigma$, but only an estimate obtained
from a sample. Robust statistical procedures to score models and test
constraints from finite samples are described at length by
\cite{sil:06}.

In the following experiments, we assume data are multivariate
Gaussian. Wishart's tetrad test can be used to evaluate $\mathcal
T(\cdot)$, which we accept as true if the p-value for the test is
greater or equal to 0.05 \citep{sil:06}.  In the {\sc
SingleLatents} procedure, for each clique $\mathcal C$ we add extra
bi-directed edges to $\mathcal G_i$ by a greedy search procedure: we
look at each pair of variables and evaluate the Bayesian information
criterion (BIC, \cite{bic:78}) for the model with the added edge. If
the best model is better than the current one, we keep the
edge. Otherwise, we stop modifying $\mathcal G_i$. An analogous
procedure is performed to add bi-directed edges in {\sc
FindMeasurementPattern}. 

In the worst-case scenario, the procedure scales at an exponential in
the number of variables due to the necessity of finding cliques in a
graph (the {\sc SingleLatents} procedure). The examples are small and
sparse enough so that this is not a problem. Some heuristics for
larger problems are described by \cite{sil:06}.

\subsection{Democratization and industrialization example}

This is the study described at the beginning of Section
\ref{sec:intro} and discussed by \cite{bol:89}. A sample of 75 countries
was collected. We will discuss the outcome of our procedure and how it
relates to the gold standard of Figure \ref{fig:bollen}.

\begin{figure}[t]
\begin{tabular}{cc}
\epsfig{file=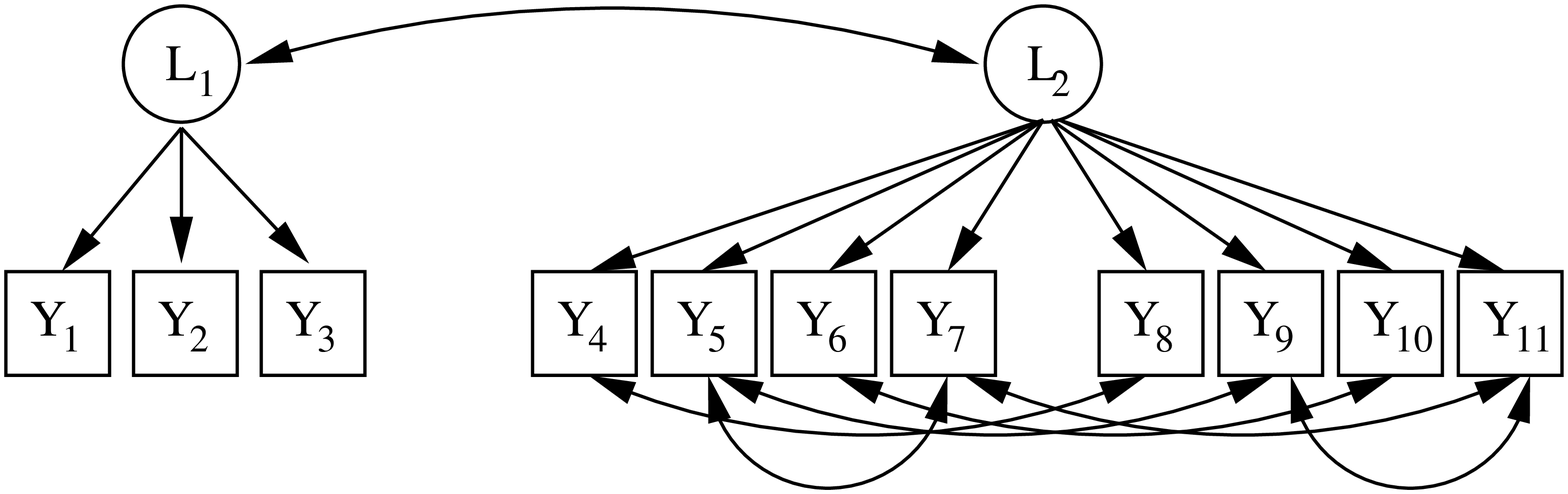,width=5.3cm}\hspace{0.3in} &
\epsfig{file=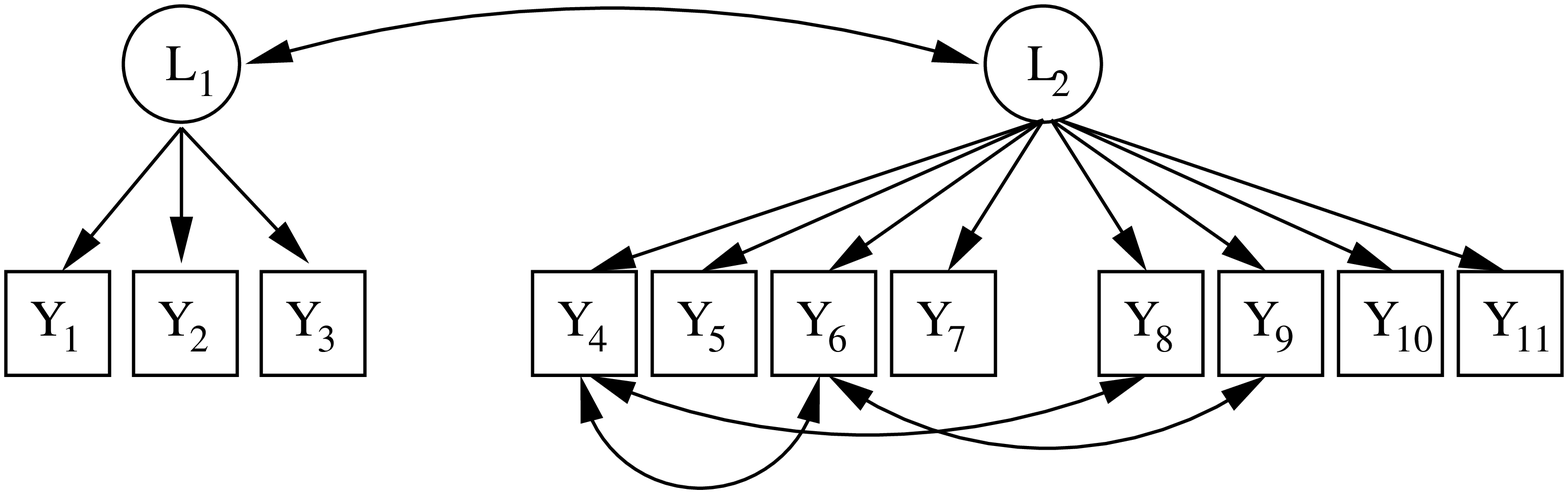,width=5.3cm}\\
(a) & (b)\\
\end{tabular}
\caption{In (a), the measurement pattern that corresponds to the
gold standard. In (b), the result of our procedure. All directed edges
are correct. With small sample sizes, the BIC score tends to produce
models simpler than expected, so it is not surprising that the model
lacks several of the bi-directed edges.}
\label{fig:bollen-result}
\end{figure}

If the true model is indeed Figure \ref{fig:bollen} and if we had
access to an oracle that could answer exactly which tetrad constraints
hold and do not hold in the true model, then the result of our
algorithm would be Figure \ref{fig:bollen-result}(a). The result
obtained with our implementation is shown in Figure
\ref{fig:bollen-result}(b). With only 75 samples, it is not surprising
that the BIC score tends to produce models with fewer edges than
expected. Still, the model reveals a lot of information present in
the expected pattern. It also suggests ways of extending the
procedure, such as allowing for the background knowledge that some
variables have the same definition, but recorded over time. Recall
that the resulting model was obtained without any extra information.

\subsection{Depression example}

The next dataset is a depression study with five indicators of
self-steem ($SELF$), four indicators of depression ($DEPRES$) and three
indicators of impulsiveness ($IMPULS$). This dataset is one of the
examples that accompany the LISREL software for structural equation
modeling. The depression data and the meaning of the corresponding
variables can also be found at 

\begin{itemize}
\item {\tt http://www.ssicentral.com/lisrel/example1-2.html}
\end{itemize}

\begin{figure}[t]
\begin{tabular}{cc}
\epsfig{file=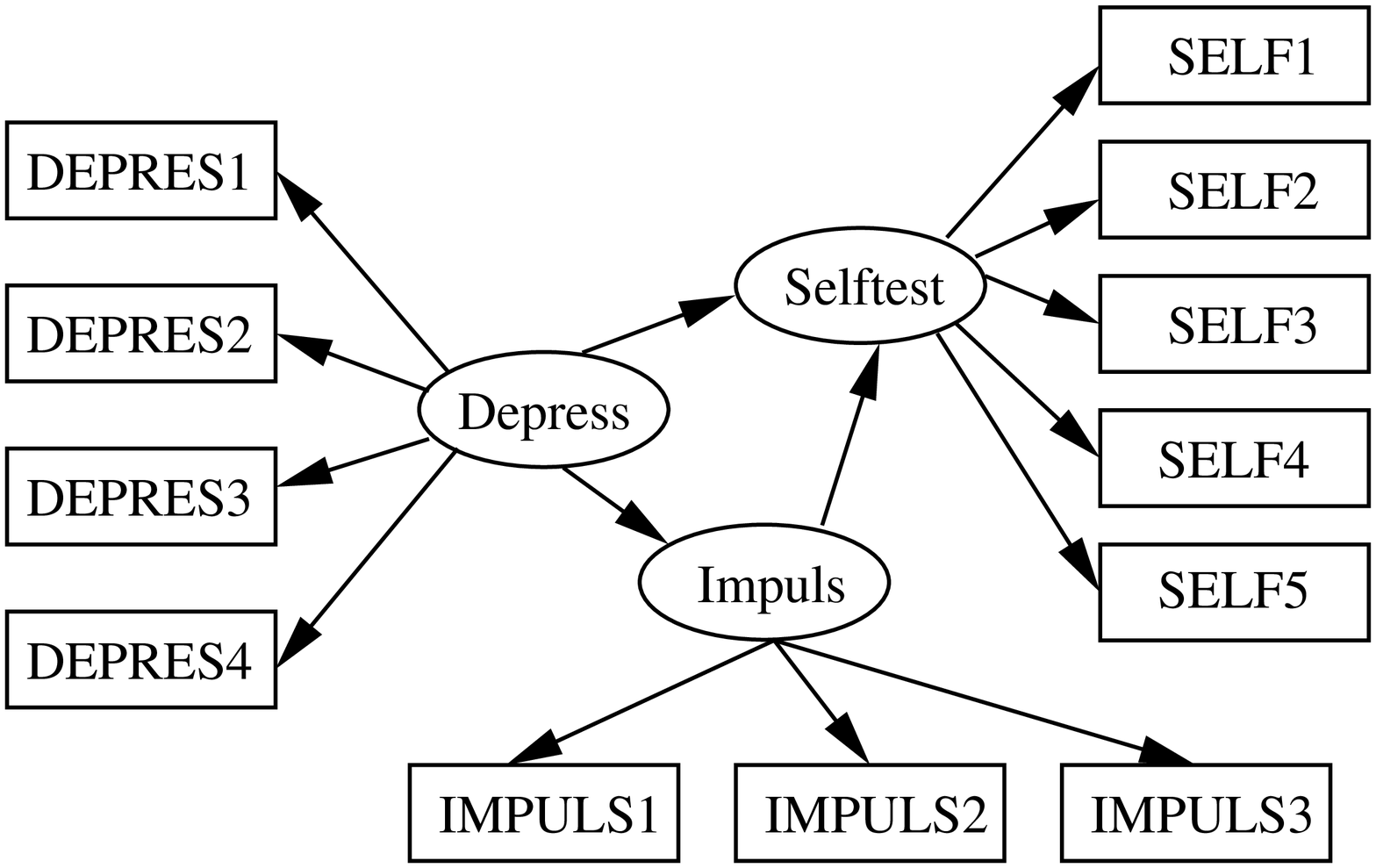,width=5.3cm}\hspace{0.3in} &
\epsfig{file=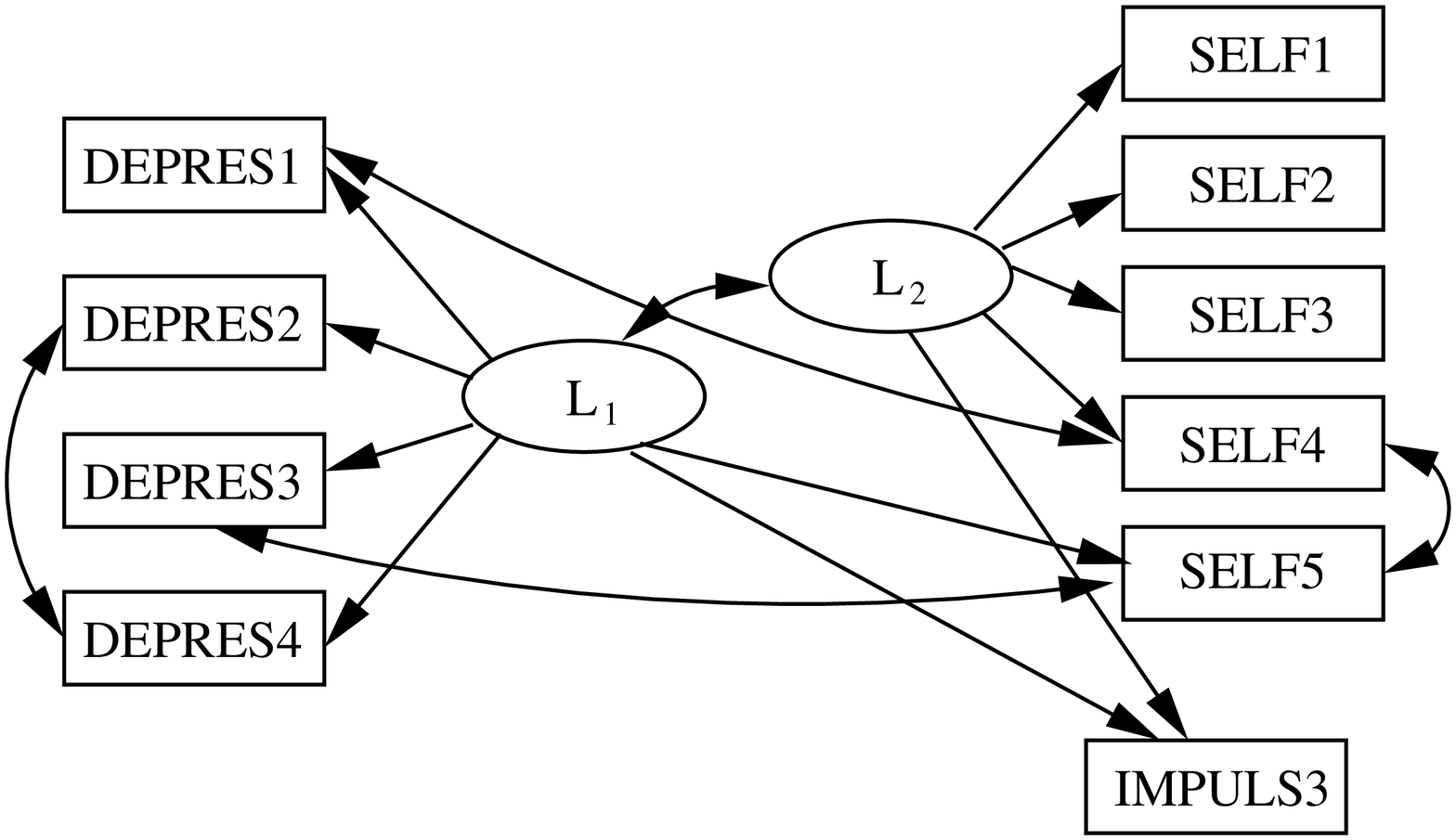,width=5.3cm}\\
(a) & (b)\\
\end{tabular}
\caption{In (a), the gold standard of the depression study. The measurement
pattern is precisely the same (except for the latent connections).
In (b), the result of our procedure. The inferred model cannot contain the
impulsiveness latent variables, as it turns out the correlation of
$IMPULS1$ and $IMPULS2$ with other variables are statistically too close to zero at a 0.10 level.}
\label{fig:depress}
\end{figure}

A theoretical gold standard is show in Figure \ref{fig:depress}(a).
It is worth mentioning that, treated as a Gaussian model, this
graphical structure does not fit the data: the chi-square score is
122.8 with 51 degrees of freedom. The sample size is 204.

Our result is shown in Figure \ref{fig:depress}(b). It was impossible
to find a hidden common cause for the indicators of impulsiveness: the
correlations of $IMPULS1$ and $IMPULS2$ with the other items were just too
low, and those items had to be discarded. The only major difference
against the gold standard was assigning $SELF5$ with the incorrect
latent parent (the role of $IMPULS3$ in the solution is compatible with
the properties of a measurement pattern). Given the number of
bi-directed connections into $SELF5$, however, this indicator seems
particularly problematic.

It is relevant to stress that in this study, the indicators are ordinal 
(in a 0 to 4 scale), not continuous. We were still able to provide relevant
information despite using a Gaussian model. In future work, methods to deal
with ordinal data will be developed. The theory for ordinal data is essentially
identical, as discussed by \cite{sil:06}. However, non-Gaussian models need
to be used, which increases the computational cost of the procedure considerably.

\section{Conclusion}
\label{sec:conclusion}

Learning measurement models is an important causal inference task in
many applied sciences. Exploratory factor analysis is a popular tool
to accomplish this task, but it can be unreliable and causal
assumptions are often left unclear. Better approaches are
needed. \cite{lohelin:04} argues that while there are several
approaches to automatically learn causal structure, none can be seem
as competitors of exploratory factor analysis. Procedures such as the
one introduced by \cite{sil:06} and extended here are important
steps that fill this gap.

The inclusion of impure indicators is an important step to make such
approaches more generally applicable. As hinted in our discussion,
other identification results to confirm or remove edges can be further
developed.  Higher-order constraints in the covariance matrix, besides
tetrad constraints, are yet to be exploited
\citep{sullivant:08}. Exploring the higher-order moments of the
observed distribution (i.e., not only the covariance matrix) has been
a successful approach to identify the causal structure of linear
models \citep{shimizu:06}, but how to adapt them to discover a
measurement model is still unclear. Finally, some progress on allowing
for non-linearities has been made \citep{sil:05b}, but more robust
statistical procedures and further identification results are
necessary.

\bibliography{arxiv_version}

\bibliographystyle{plainnat}

\end{document}